%% file: harem05.tex
\documentclass{article}
\usepackage[latin9]{inputenc}
\usepackage{geometry}
\usepackage{amsmath}
\usepackage{amssymb}
\usepackage{graphicx}

\makeatletter

\providecommand{\tabularnewline}{\\}

\date{}

\makeatother

\begin{document}

\title{Human Activity Learning and Segmentation using Partially Hidden Discriminative Models}

\author{Tran The Truyen{\small $~^{\dagger}$}, Hung H. Bui{\small $~^{\ddagger}$}
and Svetha Venkatesh{\small $~^{\dagger}$} \\
 \\
 {\small $^{\dagger}$}~Department of Computing, Curtin University
of Technology,\\
 GPO Box U 1987, Perth, Western Australia. \texttt{}~\\
\texttt{\{trantt2,svetha\}@cs.curtin.edu.au} \\
 \\
 {\small $^{\ddagger}$}~Artificial Intelligence Center, SRI International\\
 333 Ravenswood Avenue, Menlo Park, CA 94025-3493, USA. \texttt{}~\\
\texttt{bui@ai.sri.com} }
\maketitle
\begin{abstract}
Learning and understanding the typical patterns in the daily activities
and routines of people from low-level sensory data is an important
problem in many application domains such as building smart environments,
or providing intelligent assistance. Traditional approaches to this
problem typically rely on supervised learning and generative models
such as the hidden Markov models and its extensions. While activity
data can be readily acquired from pervasive sensors, e.g. in smart
environments, providing manual labels to support supervised training
is often extremely expensive. In this paper, we propose a new approach
based on semi-supervised training of partially hidden discriminative
models such as the conditional random field (CRF) and the maximum
entropy Markov model (MEMM). We show that these models allow us to
incorporate both labeled and unlabeled data for learning, and at the
same time, provide us with the flexibility and accuracy of the discriminative
framework. Our experimental results in the video surveillance domain
illustrate that these models can perform better than their generative
counterpart, the partially hidden Markov model, even when a substantial
amount of labels are unavailable.
\end{abstract}
\input{harem05_intro.tex} \input{harem05_related.tex} \input{harem05_model.tex}
\input{harem05_exp.tex} \input{harem05_con.tex}


\end{document}

%% file: harem05_intro.tex

\section{Introduction}

\label{sec:intro}


An important task in human activity recognition from low-level sensory
data is segmenting the data streams and labeling them with meaningful
sub-activities. The labels can then be used to facilitate data indexing
and organisation, to recognise higher levels of semantics, and to
provide useful context for intelligent assistive agents. The segmentation
modules are often built on top of low-level sensor components which
produce primitive and often noisy streams of events (e.g. see \cite{Ivanov:00}).
To handle the uncertainty inherent in the data, current approaches
to activity recognition typically employ probabilistic models such
as the hidden Markov models (HMMs)~\cite{RabinerIEEE-89} and more
expressive models, such as stochastic context-free grammars (SCFGs)~\cite{Ivanov:00},
hierarchical HMMs (HHMMs)~\cite{fine98hierarchical}, abstract HMMs
(AHMMs)~\cite{Bui-et-al02}, and dynamic Bayesian networks (DBNs).

All of these models are essentially generative, i.e. they model the
relation between the activity sequence $y$ and the observable data
stream $x$ via the joint distribution $p(y,x)$. Maximum likelihood
learning with these models is then performed by finding a parameter
that optimises the joint probability $p(y,x)$. This modeling approach
has two drawbacks in general. Firstly, it is often difficult to capture
complex dependencies in the observation sequence $x$, as typically,
simplifying assumptions need to be made so that the conditional distribution
$p(x|y)$ is tractable. This limits the choice of features that one
can use to encode multiple data streams. Secondly, it is often advantageous
to optimise the conditional distribution $p(y|x)$ as we do not have
to learn the data generative process. Thirdly, as we are only interested
in finding the most probable activity sequence $y^{*}=\arg\max_{y}p(y|x)$,
it is more natural to model $p(y|x)$ directly.

Thus the discriminative model $p(y|x)$ is more suitable to specify
how an activity $y$ would evolve {\em given} that we already observe
a sequence of observations $x$. In other words, the activity nodes,
rather than being the parents, become the children of the observation
nodes. With appropriate use of contextual information, the discriminative
models can represent arbitrary, dynamic long-range interdependencies
which are highly desirable for segmentation tasks.

Moreover, whilst capturing unlabeled sensor data for training is cheap,
obtaining labels in a supervised setting often requires expert knowledge
and is time consuming. In many cases we are certain about some particular
labels, for example, in surveillance data, when a person enters a
room or steps on a pressure mat. Other labels (e.g. other activities
that occur inside the room) are left unknown. Therefore, it is more
desirable to employ the semi-supervised approach. Specifically, we
consider two recent discriminative models, namely, the undirected
Conditional Random Fields (CRFs) \cite{lafferty01conditional}, (Figure
\ref{fig:models}(b)) and the directed Maximum Entropy Markov Models
(MEMMs) \cite{mcallum00maximum} (Figure \ref{fig:models}(a)). As
the original models are fully observed, we provide a treatment of
incomplete data for the CRFs and the MEMMs. The EM algorithm \cite{Dempster-et-al77}
is presented for both the models although it is not strictly required
for the CRFs.

We provide experimental results in the video surveillance domain where
we compare the performance of the proposed models and the equivalent
generative HMMs \cite{Scheffer-Wrobel01} (Figure \ref{fig:models}(c))
in learning and segmenting human indoor movement patterns. Out of
three data sets studied, a common behaviour is that the HMM is outperformed
by the discriminative counterparts even when a large portion of labels
are missing. Providing contextual features for the models increases
the performance significantly.

The novelty of this paper lies in the first work on modeling human
activity using partially hidden discriminative models. Although semi-supervised
learning has been investigated for a while, much work has concentrated
on unstructured data and classification. There has been little work
on structured data and segmentation and how much labeling effort are
needed.

The remainder of the paper is organised as follows. Section \ref{sec:related}
reviews related work in human activity segmentation and background
in CRFs and MEMMs and in semi-supervision. Section \ref{sec:CRFs}
describes the partially hidden discriminative models. The paper then
describes implementation and experiments and presents results in Section
\ref{sec:experiments}. The final section summarises major findings
and further work. 

%% file: harem05_related.tex

\section{Related work}

\label{sec:related}

Hidden Markov models (HMMs) have been used to model simple human activities
and human motion patterns~\cite{GRZEGORZ_EL-03,aggarwal99human,Yamato-et-alCVPR92}.
More recent approaches have used more sophisticated generative models
to capture the hierarchical structure of complex activities. The abstract
hidden Markov model (AHMM)~\cite{Bui-et-al02} is used in~\cite{liao2007learning}
to model human transportation patterns from outdoor GPS sensors, and
in~\cite{OSENTOSKI_EL-04} to model human indoor motion patterns
from sensors placed in mobile robots. Using the AHMM, multiple levels
of semantics can be built on top of the HMMs allowing flexibility
in modeling the evolution of activities across multiple levels of
abstraction. To learn the parameters, the expectation maximisation
(EM) algorithm can be used. However, these models are generative,
and are not suitable to work with arbitrary or overlapping features
in the data streams.

Discriminative models specify the conditional probability $p(y|x)$
without modeling the data $x$. Let $y=\{y_{i:n}\}$ and assume that
the probability $p(y|x)$ is specified with respect to a graph $\mathcal{G}=(\mathcal{E},\mathcal{V})$,
where each vertex $i\in\mathcal{V}$ represents a random variable
$y_{i}$ and the edges $e\in\mathcal{E}$ encode the correlation between
variables. The graph $\mathcal{G}$ can be undirected, as in the Conditional
Random Fields (CRFs) \cite{lafferty01conditional} (Figure \ref{fig:models}(b))
or directed as in the Maximum Entropy Markov Models (MEMMs) \cite{mcallum00maximum}
(Figure \ref{fig:models}(a)). The CRFs define the model as follows
\begin{eqnarray}
p(y|x;\lambda)=\frac{1}{Z(x;\lambda)}\prod_{c}\Psi_{c}(y_{c},x;\lambda)\label{Hammersley-Clifford}
\end{eqnarray}
where $c$ is the clique defined by the structure of $\mathcal{G}$,
$\Psi_{c}(y_{c},x;\lambda)$ is the potential function defined over
the clique $c$, $\lambda$ are model parameters, and $Z(x;\lambda)=\sum_{y}\prod_{c}\Psi_{c}(y_{c},x;\lambda)$
is the normalisation factor.

We consider the chain structure CRFs for our labeling tasks (Figure
\ref{fig:models}(b)), that is $y=\{y_{1:T}\}$. The potential function
becomes $\Psi_{t}(y_{t-1},y_{t},x;\lambda)$, which is then typically
parameterised using the log-linear model $\Psi_{t}(y_{t-1},y_{t},x;\lambda)=\exp(\sum_{k}\lambda_{k}f_{k}(y_{t-1},y_{t},x)$.
The functions $\{f_{k}(y_{t-1},y_{t},x)\}$ are the features that
capture the statistics of the data and the semantics at time $t$.
The parameters $\lambda$ are the weight associated with the features
and are estimated through training.

The MEMM is a directed, local version of the CRFs (Figure \ref{fig:models}(a)),
in which each source state $j$ has a conditional distribution 
\begin{eqnarray}
p_{j}(y_{t}|x_{t};\lambda)=p(y_{t}|y_{t-1}=j,x_{t};\lambda)=\frac{1}{Z(x_{t},j)}\exp(\sum_{k}\lambda_{jk}f_{k}(x_{t},y_{t}))\label{MEMMs-model}
\end{eqnarray}
where $\lambda_{jk}$ are parameters of the source state $y_{t-1}=j$.
The MEMMs can also be considered as conditionally trained HMMs (e.g.
see the difference between Figures \ref{fig:models}(a,c)). Although
CRFs solve the {\em label bias} problem associated with the local
normalised MEMMs \cite{lafferty01conditional}, we believe that the
MEMMs are useful in learning and understanding activity patterns because
they directly encode the temporal state evolution through the transition
model $p(y_{t}|y_{t-1}=j,x_{t};\lambda)$.

Supervised learning in the CRFs and MEMMs typically maximises the
conditional log-likelihood %
\footnote{For multiple iid data instances, we should write $\mathcal{L}(\lambda)=\sum_{x}\tilde{p}(x)\log p(y|x;\lambda)$
where $\tilde{p}(x)$ is the empirical distribution of training data,
but we drop this notation for clarity.%
} $\mathcal{L}(\lambda)=\log p(y|x;\lambda)$. Gradient-based methods
\cite{sha-pereira:2003:HLTNAACL} are considered the fastest up to
now.

Partially hidden models have received significant attention recently.
The partially hidden Markov model (PHMM) proposed in \cite{Scheffer-Wrobel01}
(Figure \ref{fig:models}(c)) addresses the similar partial labeling
problem as ours and we will use this model to compare with our discriminative
models. In \cite{NIPS2005_810}, CRFs with a hidden layer are introduced
but labels are never given for this layer, thus they are not concerned
with how robust the model is with respect to amount of missing data.
The idea of {\em constrained} inference is introduced in \cite{Kristjannson-et-al04}
but they do not address the learning problem as we do. The more recent
work in \cite{Culotta-McCallum-AAAI05} extends the work of \cite{Kristjannson-et-al04}
to learning and addresses the interactive labeling effort by users.
The results, however, are difficult to generalise to non-interactive
applications in a non-active learning fashion. 

%% file: harem05_model.tex

\section{Partially hidden discriminative models}

\label{sec:CRFs}

\subsection{The models}

\begin{figure}[htb]
\begin{centering}
\begin{tabular}{ccccc}
\includegraphics[width=0.27\linewidth]{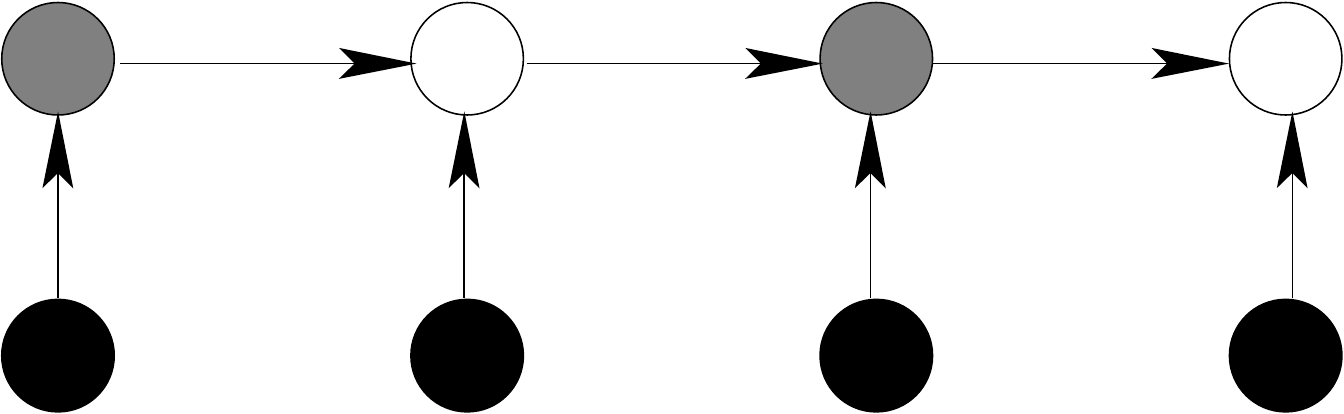}  &  & \includegraphics[width=0.31\linewidth]{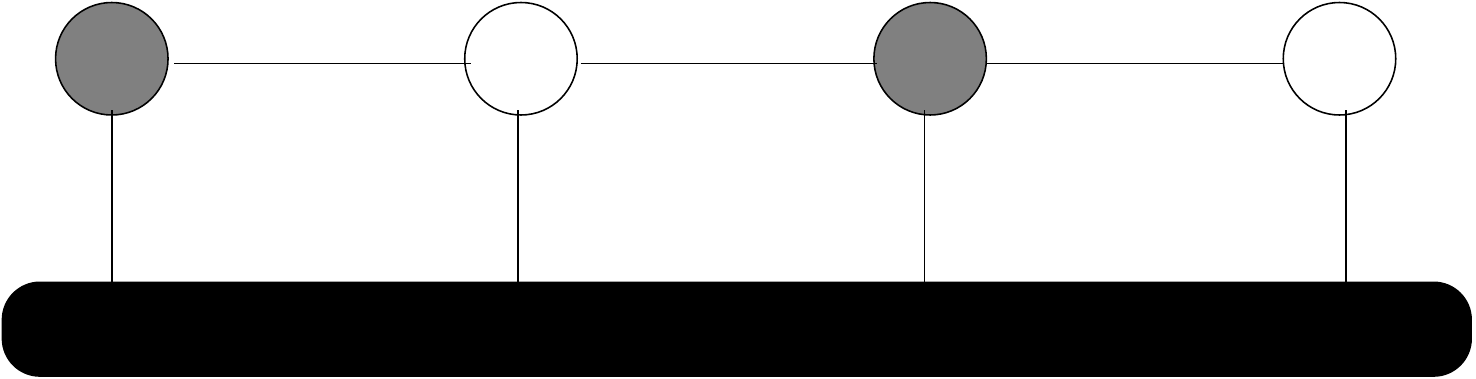}  &  & \includegraphics[width=0.27\linewidth]{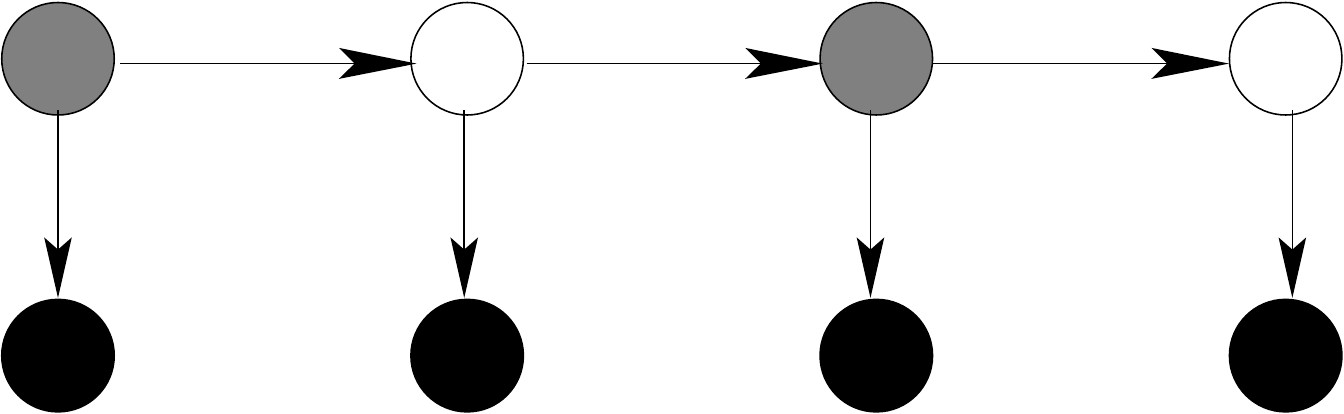} \tabularnewline
(a) MEMM  &  & (b) CRF  &  & (c) PHMM \tabularnewline
\end{tabular}
\par\end{centering}

\caption{(a,b): The partially hidden discriminative models, and (c): partially
hidden Markov models. Filled circles and bars are data observations,
empty circles are hidden labels, shaded circles are the visible labels}

\label{fig:models} 
\end{figure}

In our partially hidden discriminative models, the label sequence
$y$ consists of a visible component $v$ (e.g. labels that are provided
manually, or are acquired automatically by reliable sensors) and a
hidden part $h$ (labels that are left unspecified or those we are
unsure). The joint distribution of all visible variables $v$ is therefore
given as 
\begin{eqnarray}
p(v|x;\lambda)=\sum_{h}p(v,h|x;\lambda)=\sum_{h}p(y|x;\lambda)
\end{eqnarray}

\textbf{CRFs}. For the log-linear CRFs, we have 
\begin{eqnarray}
p(y|x;\lambda)=\frac{1}{Z(x)}\prod_{t}\exp(\sum_{k}\lambda_{k}f_{k}(y_{t-1},y_{t},x))
\end{eqnarray}
where $Z(x)=\sum_{y}\prod_{t}\exp(\sum_{k}\lambda_{k}f_{k}(y_{t-1},y_{t},x))$.
In this case, the complexity of computing $p(v|x;\lambda)$ is the
same as that of computing the partition function $Z(x)$ up to a constant.
Note that $Z(x)$ has the sum-product form, which can be computed
efficiently using a single forward pass.

\textbf{MEMMs}. As stated in Section \ref{sec:related}, directed
models like the MEMMs are important in activity modeling because they
naturally encode the state transitions given the observations. Here
we offer a slightly more general view of the MEMMs in that we define
a single model for all source states rather than separate models for
each source state as in (\ref{MEMMs-model}). In addition, as the
model is discriminative, we do not have to model the observation sequence
$x$. Thus we are free to encode arbitrary information exacted from
the whole sequence $x$ to the local distribution. In our implementation,
this is realised by using a sliding window of size $s$ centred at
the current time $t$ to capture the local context of the observation.
The local distribution reads 
\begin{eqnarray}
p(y_{t}|\Omega_{t},y_{t-1};\lambda)=\frac{1}{Z(\Omega_{t},y_{t-1})}\exp(\sum_{k}\lambda_{k}f_{k}(\Omega_{t},y_{t-1},y_{t}))\label{GMEMMs-model}
\end{eqnarray}
where $\Omega_{t}=\{x_{(t-s_{1}):(t+s_{2})}\}$ is the context of
size $s=s_{1}+s_{2}+1$, and the parameter set $\{\lambda_{k}\}$
is now shared across the states. This view of MEMMs reduces to the
original model if the feature set $\{f_{k}(\Omega_{t},y_{t-1},y_{t})\}$
consists of only indicator functions of states. The new view thus
enjoys the same probabilistic inference properties but the learning
is slightly different from the MEMM as it incorporates the structural
constraint via the shared parameters while the MEMMs learns each local
classifiers independently. The use of contextual features reflects
the fact that the the current activity $y_{t}$ is generally correlated
with the past and the future of sensor data.

As the graphical model of the MEMMs forms a Markov chain conditioned
on the observation $x$, the joint incomplete distribution is therefore
\begin{eqnarray}
p(v|x;\lambda)=\sum_{h}\prod_{t}p(y_{t}|\Omega_{t},y_{t-1};\lambda)
\end{eqnarray}
Again, this is a sum-product case, which can be computed by a single
forward pass.

\subsection{Parameters learning}

To learn the model parameters that are best explained by the data,
we maximise the penalised log-likelihood 
\begin{eqnarray}
\Lambda(\lambda)=\mathcal{L}(\lambda)-\frac{1}{2\sigma^{2}}||\lambda||^{2}\label{eq:obj-func}
\end{eqnarray}
where $\mathcal{L}(\lambda)=\log p(v|x;\lambda)$. The regularisation
term is needed to avoid over-fitting when only limited data is available
for training. For simplicity, the parameter $\sigma$ is shared among
all dimensions and is selected experimentally.

As with incomplete data, an alternative to maximise the log-likelihood
is using the EM algorithm \cite{Dempster-et-al77} whose Expectation
(E-step) is to calculate the quantity 
\begin{eqnarray}
Q(\lambda^{j},\lambda)=\sum_{h}p(h|v,x;\lambda^{j})\log p(h,v|x)\label{EM}
\end{eqnarray}
and the Maximisation (M-step) maximises the concave lower bound of
the log-likelihood $Q(\lambda^{j},\lambda)-\frac{1}{2\sigma^{2}}||\lambda||^{2}$
with respect to $\lambda$. Unlike Bayesian networks, the log-linear
models do not yield closed form solutions in the the M-step. However,
as the function $Q(\lambda^{j},\lambda)$ is concave, it is still
advantageous to optimise with efficient Newton-like algorithms.

\textbf{CRFs}. For the partially hidden CRFs, the gradient of incomplete
likelihood reads 
\begin{eqnarray}
\frac{\partial\mathcal{L}(\lambda)}{\partial\lambda_{k}}=\sum_{t}\sum_{h_{t-1},h_{t}}p(h_{t-1},h_{t}|v,x;\lambda)f_{k}(h_{t-1},h_{t},v,x)-\sum_{t}\sum_{y_{t-1},y_{t}}p(y_{t-1},y_{t}|x;\lambda)f_{k}(y_{t-1},y_{t},x)\label{CRF-ll-grad}
\end{eqnarray}
Zeroing the gradient does not yield an analytical solution, so typically
iterative numerical methods such as conjugate gradient and Newton
methods are needed. The gradient of the lower bound in the EM framework
of (\ref{EM}) is similar to (\ref{CRF-ll-grad}), except that the
pairwise marginals $p(h_{t-1},h_{t}|v,x;\lambda)$ are now replaced
by the marginals of the previous EM iteration $p(h_{t-1},h_{t}|v,x;\lambda^{j})$.
The pairwise marginals $p(y_{t-1},y_{t}|x)$ can be computed easily
using a forward pass and a backward pass in the standard message passing
scheme on the chain. Details are omitted for space constraint.

\textbf{MEMMs}. In learning of MEMMs, the E-step is to calculate 
\begin{eqnarray}
Q(\lambda^{j},\lambda)=\sum_{t}\sum_{h_{t-1}}p(h_{t-1}|v,\Omega_{t};\lambda^{j})\sum_{h_{t}}p(h_{t}|h_{t-1},\Omega_{t};\lambda^{j})\log p(h_{t}|h_{t-1},\Omega_{t};\lambda)\label{MEMM-E}
\end{eqnarray}
and the M-step is to solve the zeroing gradient equation 
\begin{eqnarray*}
\frac{\partial Q(\lambda^{j},\lambda)}{\partial\lambda_{k}}\left\{ \right\}  & = & \sum_{t}\sum_{h_{t-1}}p(h_{t-1}|v,\Omega_{t};\lambda^{j})A_{t}(h_{t-1})\,\,\mbox{where}\\
A_{t}(h_{t-1}) & = & \sum_{h_{t}}p(h_{t}|h_{t-1},\Omega_{t};\lambda^{j})f_{k}(h_{t-1},h_{t},\Omega_{t})-\sum_{y_{t}}p(y_{t}|h_{t-1},\Omega_{t};\lambda)f_{k}(h_{t-1},y_{t},\Omega_{t})
\end{eqnarray*}
Computation of the EM reduces to that of marginals and state transition
probabilities, which can be carried out efficiently in the Markov
chain framework using dynamic programming.

\subsection{Segmentation}

For segmentation, we use the MAP assignment $y^{*}=\arg\max_{y}p(y|x,\lambda)$
to infer the most probable label sequence $y^{*}$ for a given data
sequence $x$. For both the CRFs and MEMMs, the Viterbi algorithm
\cite{RabinerIEEE-89} can be naturally adapted. If some labels are
provided (e.g. by some reliable sensors, or by users in interactive
applications) we have the so-called {\em constrained} inference
\cite{Kristjannson-et-al04}, but this is a trivial adaptation of
the Viterbi decoding \cite{RabinerIEEE-89}.

\subsection{Comparison with the PHMMs}

The main difference between the models described in this section (Figure
\ref{fig:models}(a,b)) and the PHMMs \cite{Scheffer-Wrobel01} (Figure
\ref{fig:models}(c)) is the conditional distribution $p(y|x)$ in
discriminative models compared to the joint distribution $p(y,x)$
in the PHMMs. The data distribution of $p(x)$ and how $x$ is generated
are not of concern in the discriminative models. In the PHMMs, on
the contrary, the observation point $x_{t}$ is presumably generated
by the parent label node $y_{t}$, so care must be taken to ensure
proper conditional independence among $\{x_{t}\}_{t=1}^{T}$. This
difference has an implication that, while the discriminative models
may be good to encode the output labels directly with arbitrary information
extracted from the whole observation sequence $x$, the PHMMs better
represent $x$ when little information is associated with $y$. For
example, when $y$ is totally missing, $p(x)=\sum_{y}p(y,x)$ is still
modeled in the PHMMs and provides useful information. Our experiments
in the next section show this difference more clearly.

Moreover, whilst we employ the log-linear models with unconstrained
parameters, the PHMMs use the constrained transition and emission
probabilities as parameters. In terms of modeling label `visibility',
the PHMMs are more general as they allow a subset of labels to be
associated with certain nodes, and not only a full set as in hidden
nodes or a single label as in visible nodes. However, it is quite
straightforward to extend our partially hidden discriminative models
to incorporate the same representation.

%% file: harem05_exp.tex

\section{Experiments and results}

\label{sec:experiments}

Our task is to infer the activity patterns of a person (the actor)
in a video surveillance scene. The observation data is provided by
static cameras while the labels, which are activities such as {\em
`go-from-A-to-B'} during the time interval $[t_{a},t_{b}]$ (see
Table~\ref{activities}), are recognised by the trained models.


\subsection{Setup and data}

\begin{figure}[htb]
\begin{centering}
\begin{tabular}{ccc}
\includegraphics[width=0.4\linewidth]{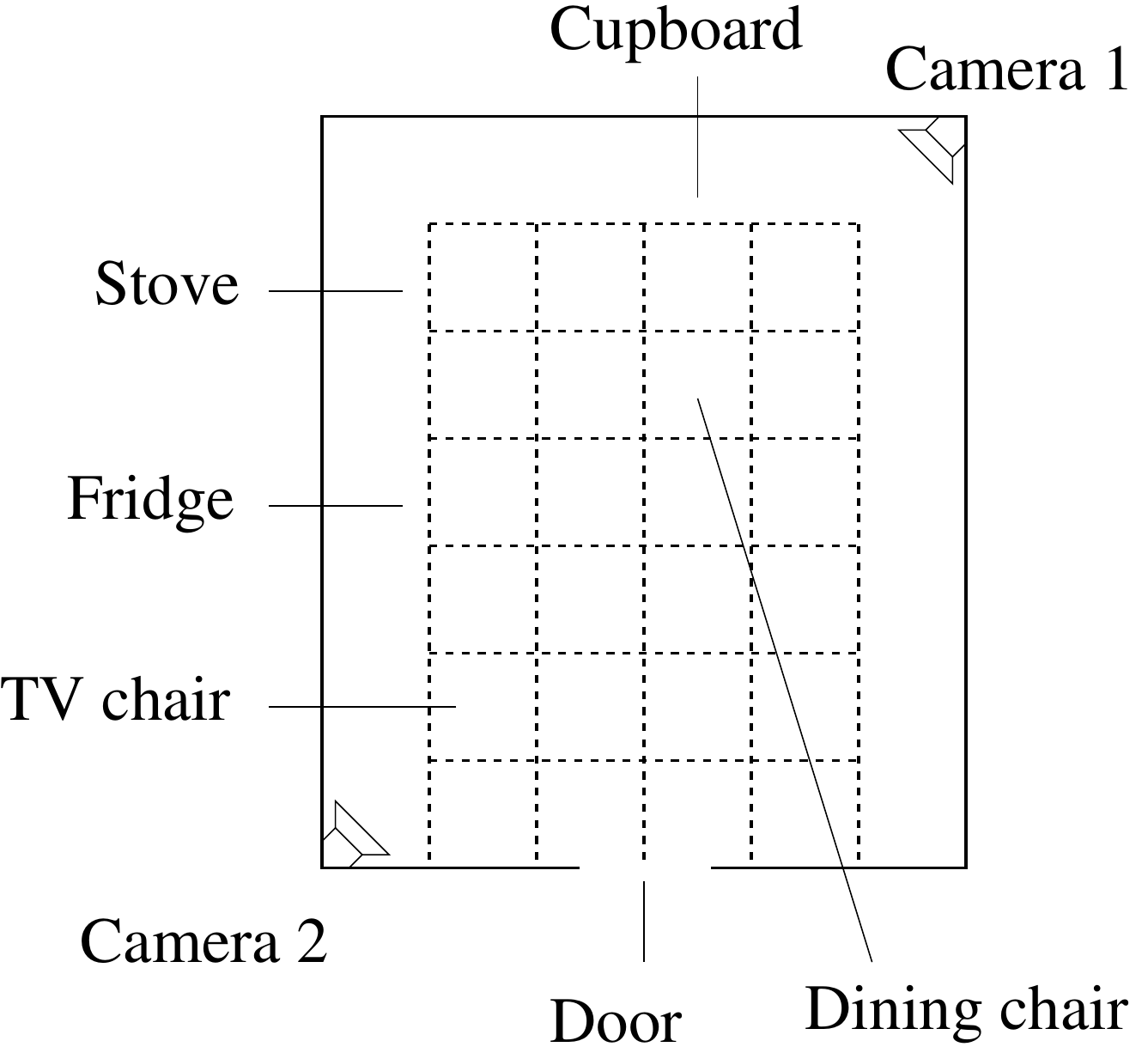}  & \includegraphics[width=0.21\linewidth]{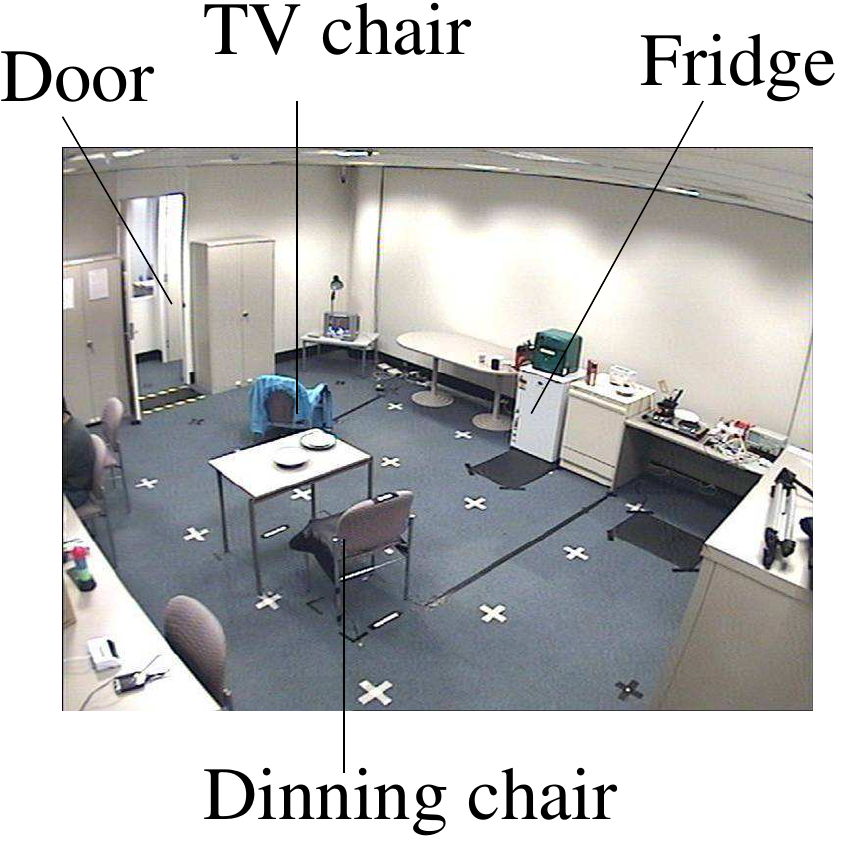} & \includegraphics[width=0.21\linewidth]{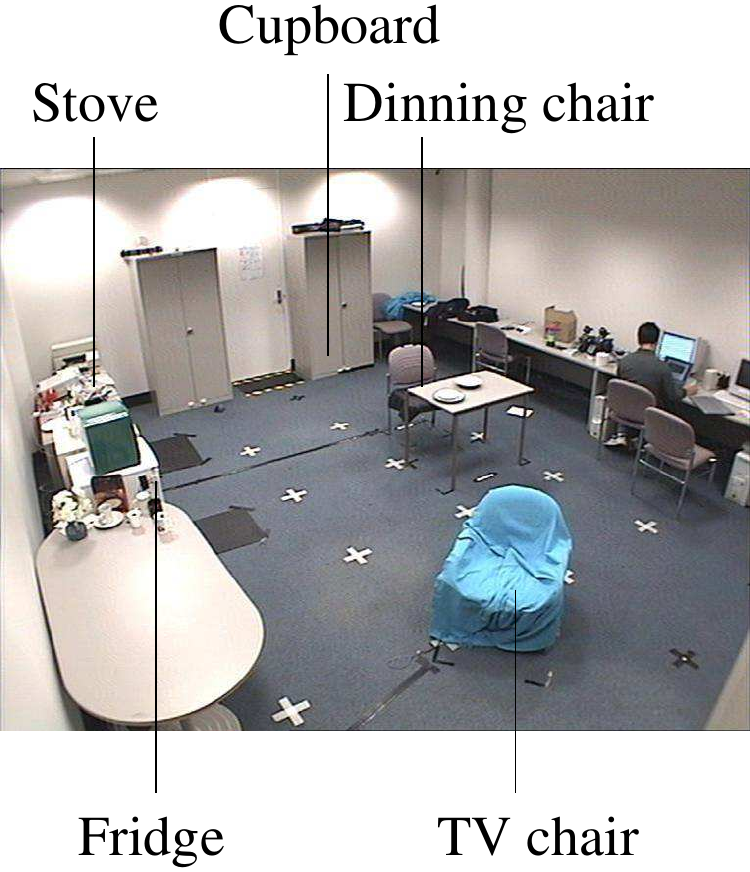}\tabularnewline
\end{tabular}
\par\end{centering}

\caption{The environment and scene viewed from the two cameras.}

\label{fig:cam} 
\end{figure}

The surveillance environment is a $4\times6m^{2}$ dining room and
kitchen (Figure~\ref{fig:cam}). Two static cameras are installed
to capture the video of the actor making some meals. There are six
landmarks which the person can visit during the meals: door, TV chair,
fridge, stove, cupboard, and dining chair. Figure~\ref{fig:cam}
shows the room and the special landmarks viewed from the two cameras.

\begin{table}
\caption{The primitive activities (the labels).}

\begin{centering}
\begin{tabular}{|c|l||c|l|}
\hline 
Activity  & Landmarks  & Activity  & Landmarks\tabularnewline
\hline 
\hline 
1  & Door$\rightarrow$Cupboard  & 7  & Fridge$\rightarrow$TV chair\tabularnewline
\hline 
2  & Cupboard$\rightarrow$Fridge  & 8  & TV chair$\rightarrow$Door\tabularnewline
\hline 
3  & Fridge$\rightarrow$Dining chair  & 9  & Fridge$\rightarrow$Stove\tabularnewline
\hline 
4  & Dining chair$\rightarrow$Door  & 10  & Stove$\rightarrow$Dining chair\tabularnewline
\hline 
5  & Door$\rightarrow$TV chair  & 11  & Fridge$\rightarrow$Door\tabularnewline
\hline 
6  & TV chair$\rightarrow$Cupboard  & 12  & Dining chair$\rightarrow$Fridge\tabularnewline
\hline 
\end{tabular}
\par\end{centering}

\label{activities} 
\end{table}

\begin{figure}[htb]
\begin{centering}
\begin{tabular}{c}
\includegraphics[width=0.45\linewidth]{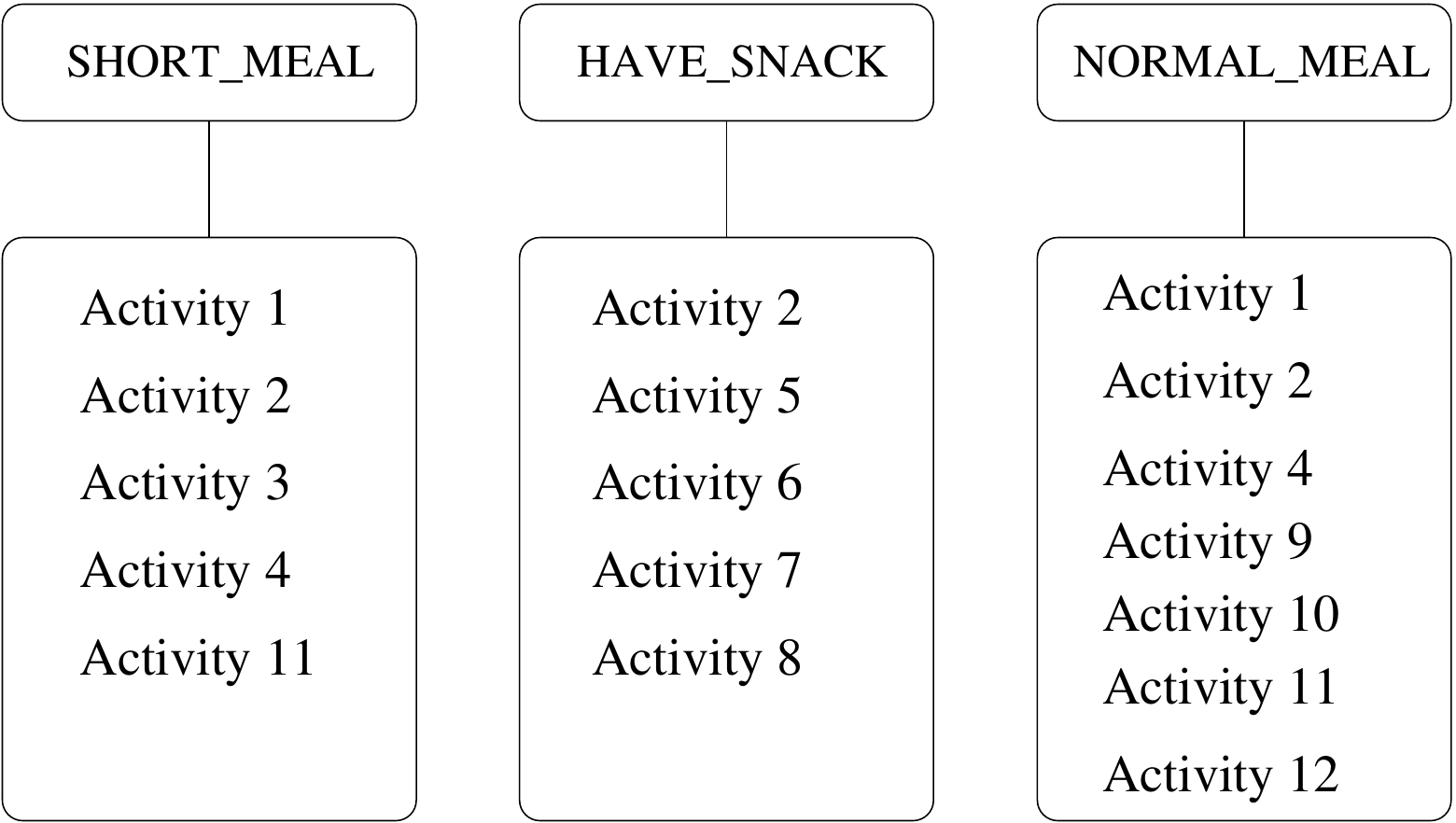} \tabularnewline
\end{tabular}
\par\end{centering}

\caption{Associated primitive activities.}

\label{fig:beh} 
\end{figure}

We study three scenarios corresponding to the person making a short
meal (denoted by SHORT\_MEAL), having a snack (HAVE\_SNACK), and making
a normal meal (NORMAL\_MEAL). Each scenario comprises of a number
of primitive activities as listed in Table \ref{activities}. Figure~\ref{fig:beh}
shows the association between scenarios and their primitive activities.
The SHORT\_MEAL data set has 12 training and 22 testing video sequences;
and each of the HAVE\_SNACK and NORMAL\_MEAL data sets consists of
15 training and 11 testing video sequences. For each raw video sequence
captured, we use a background subtraction algorithm to extract a corresponding
discrete sequence of coordinates of the person based on the person's
bounding box. The training sequences are partially labeled, indicated
by the portion of missing labels $\rho$. The testing sequences provide
the ground-truth for the algorithms. The sequence length ranges from
$T=20-60$ and the number of labels per sequence is allowed to vary
as $T*(1-\rho)$ where $\rho\in[0,100\%]$.

We apply standard evaluation metrics such as precision $P$, recall
$R$, and the $F1$ score given as $F1=2*P*R/(P+R)$ on a per-token
basis.


\subsection{Feature design and contextual extraction}

\label{sec:feature-design} Features are crucial components of the
model as they tie raw observation data with semantic outputs (i.e.
the labels). The features need to be discriminative enough to be useful,
and at the same time, they should be as simple and intuitive as possible
to reduce manual labour. The current raw data extracted from the video
contains only $(X,Y)$ coordinates. From each coordinate sequences,
at each time slice $t$, we extract a vector of five elements from
the observation sequence $g(x,t)=(X_{t},Y_{t},u_{X_{t}},u_{Y_{t}},s_{t}=\sqrt{u_{X_{t}}^{2}+u_{Y_{t}}^{2}})$,
which correspond to the $(X,Y)$ coordinates, the $X$ \& $Y$ velocities,
and the speed, respectively. Since the extracted coordinates are fairly
noisy, we use the average velocity measurement within a time interval
of small width $w$, i.e. $u_{X_{t}}=(X_{t+w/2}-X_{t-w/2})/w$. Typically,
these observation-based features are real numbers and are normalised
so that they have a similar scale.

We decompose the feature set $\{f_{k}(y_{t-1},y_{t},x)\}$ into two
subsets: the {\em state-observation} features 
\begin{eqnarray}
f_{l,m,\epsilon}(x,y_{t}):=\mathbb{I}[y_{t}=l]h_{m}(x,t,\epsilon)\label{data-ass-feature}
\end{eqnarray}
and the {\em state-transition} features 
\begin{eqnarray}
f_{l_{1},l_{2}}(y_{t-1},y_{t}):=\mathbb{I}[y_{t-1}=l_{1}]\mathbb{I}[y_{t}=l_{2}]
\end{eqnarray}
where $m=1..5$ and $h_{m}(x,t,\epsilon)=g_{m}(x,t+\epsilon)$ with
$\epsilon=-s_{1},..0,..s_{2}$ for some positive integers $s_{1}$,
$s_{2}$. The state-observation features in (\ref{data-ass-feature})
thus incorporate neighbouring observation points within a sliding
window of width $s=s_{1}+s_{2}+1$. This is intended to capture the
correlation of the current activity with past and future observations,
and is a realisation of the temporal \textit{context} $\Omega_{t}$
of the observations in (\ref{GMEMMs-model}). Thus the feature set
has $K=5s|Y|+|Y|^{2}$ features, where $|Y|$ is the number of distinct
label symbols.

\begin{figure}[htb]
\begin{centering}
\begin{tabular}{ccc}
\includegraphics[width=0.3\linewidth]{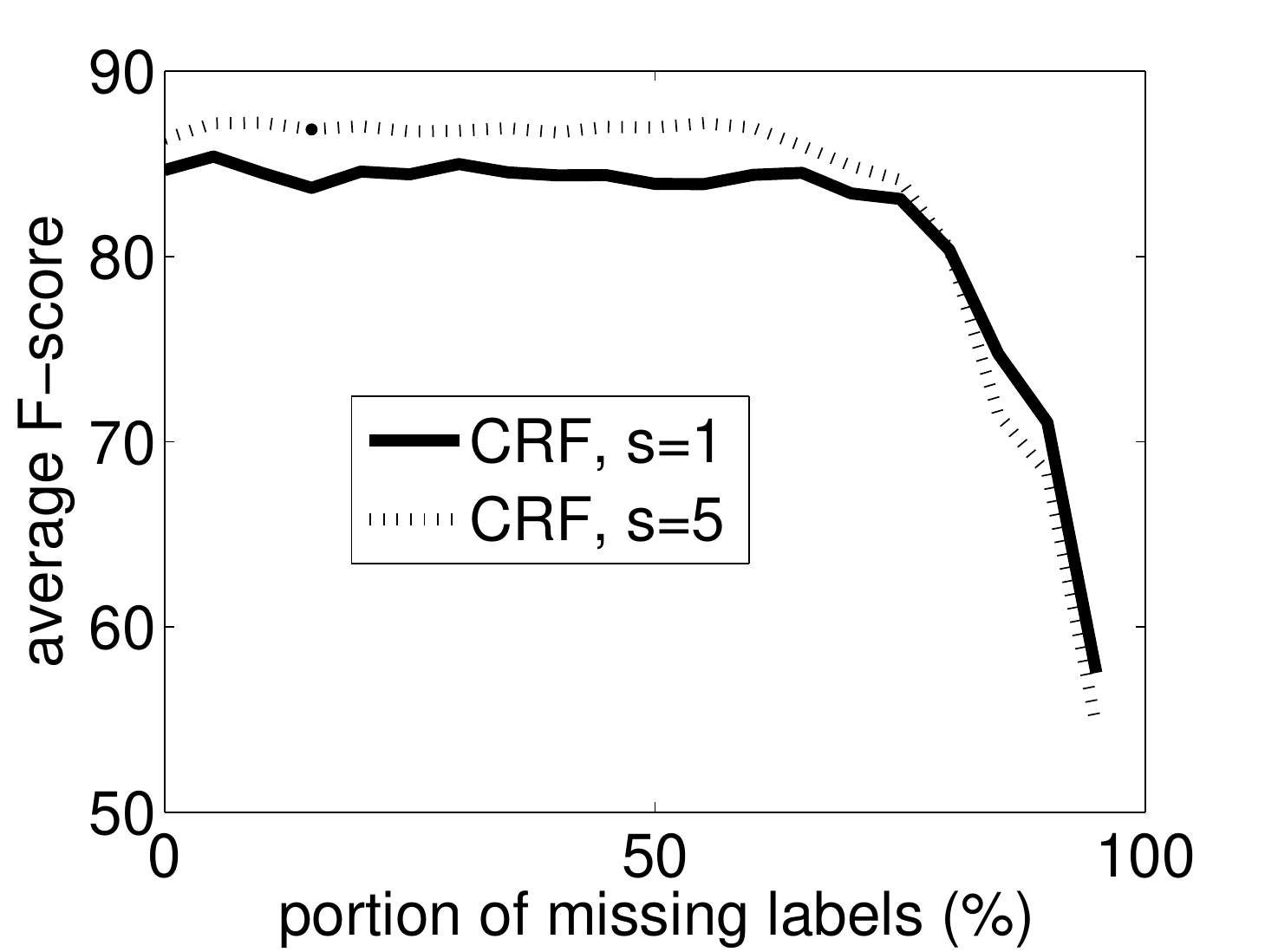}  &  & \includegraphics[width=0.3\linewidth]{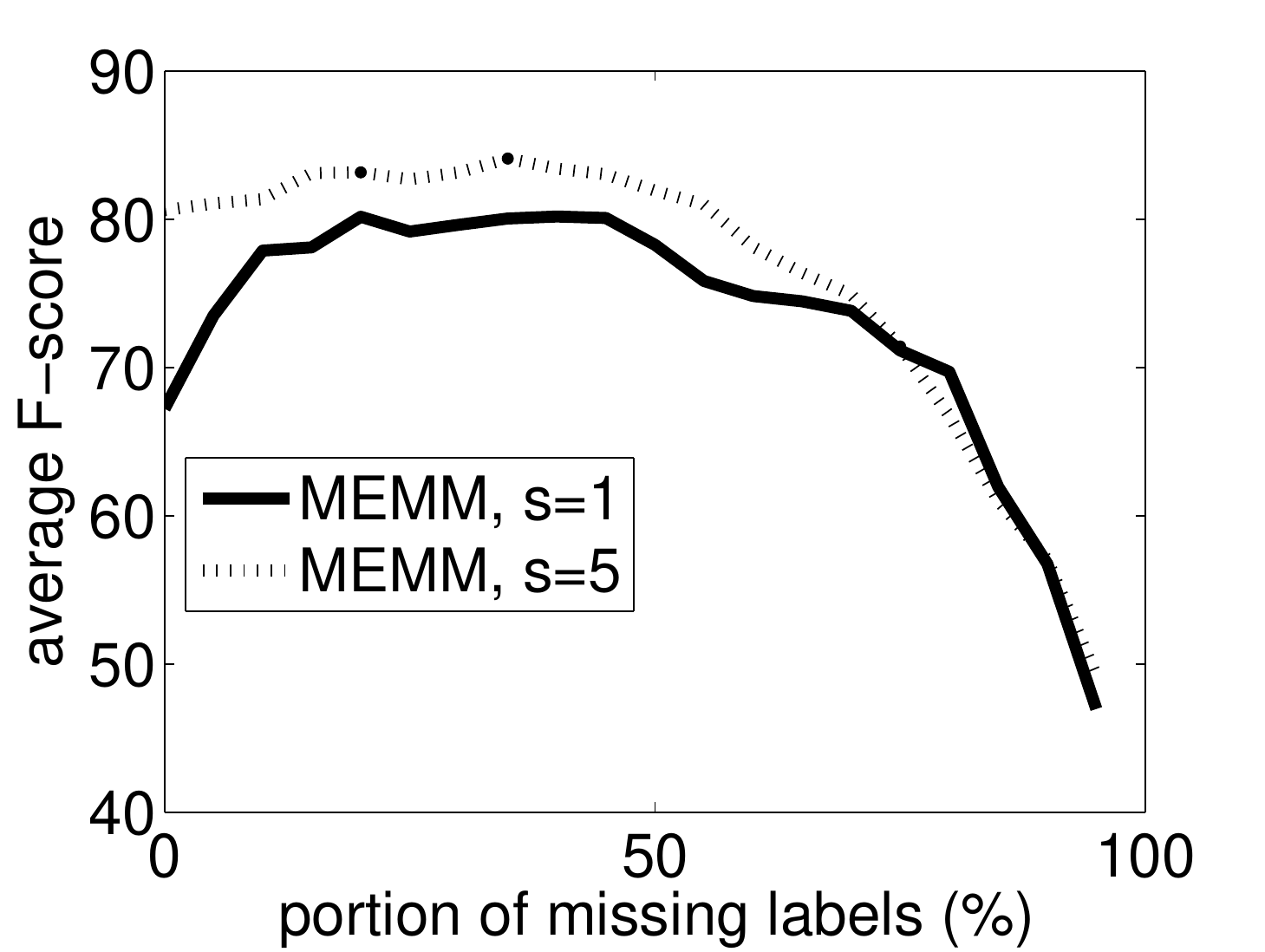}\tabularnewline
(a)  &  & (b) \tabularnewline
\end{tabular}\caption{The role of context (SHORT\_MEAL), $s$: the window size to extract
observation data. (a) CRFs, (b) MEMMs. In all figures, the x-axis:
the portion of missing labels (\%) and the y-axis: the averaged F-score
(\%) over all states and over 10 repetitions.}

\par\end{centering}

\centering{}\label{fig:feature-training} 
\end{figure}

To have a rough idea of how the observation context influences the
performance of the models, we try different window sizes $s$ (see
Equation (\ref{MEMMs-model})). The experiments show that incorporating
the context of observation sequences does help to improve the performance
significantly (see Figure \ref{fig:feature-training}). We did not
try exhaustive searches for the best context size, nor did we implement
any feature selection mechanisms. As the number of features scales
linearly with the context size as $K=5s|Y|+|Y|^{2}$, where $s$ can
be any integer between 1 and $T$, where $T$ is the sequence length,
clearly a feature selection algorithm is needed when we want to capture
long range correlation. For the practical purposes of this paper,
we choose $s=5$ for both CRFs and MEMMs. Thus in our experiments,
CRFs and MEMMs share the same feature set, making the comparison between
the two models consistent.


\subsection{Performance of models}

To evaluate the performance of discriminative models against the equivalent
generative counterparts, we implement the PHMMs (Figure \ref{fig:models}(c)).
The features extracted from the sensor data for the PHMMs include
the discretised position and velocity. These features are different
from those used in discriminative models in that discriminative features
can be continuous.

To train discriminative models, we implement the non-linear conjugate
gradient (CG) of Polak-Ribière and the limited memory quasi-Newton
L-BFGS. After several pilot runs, we select the L-BFGS to optimise
the objective function in Eq.~(\ref{eq:obj-func}) directly. In the
case of MEMMs, the regularised EM algorithm is chosen together with
the CG. The algorithms stop when the rate of convergence is less than
$10^{-5}$. The regularisation constants are empirically selected
as $\sigma=5$ in the case of CRFs, and $\sigma=20$ in the case of
MEMMs.

For the PHMMs, it is observed that the initial parameter initialisation
is critical to learn the correct model. Random initialisations often
result in very poor performance. This is unlike the discriminative
counterparts in which all initial parameters can be trivially set
to zeros (equally important).

\begin{figure}[htb]
\begin{centering}
\begin{tabular}{ccc}
\includegraphics[width=0.3\linewidth]{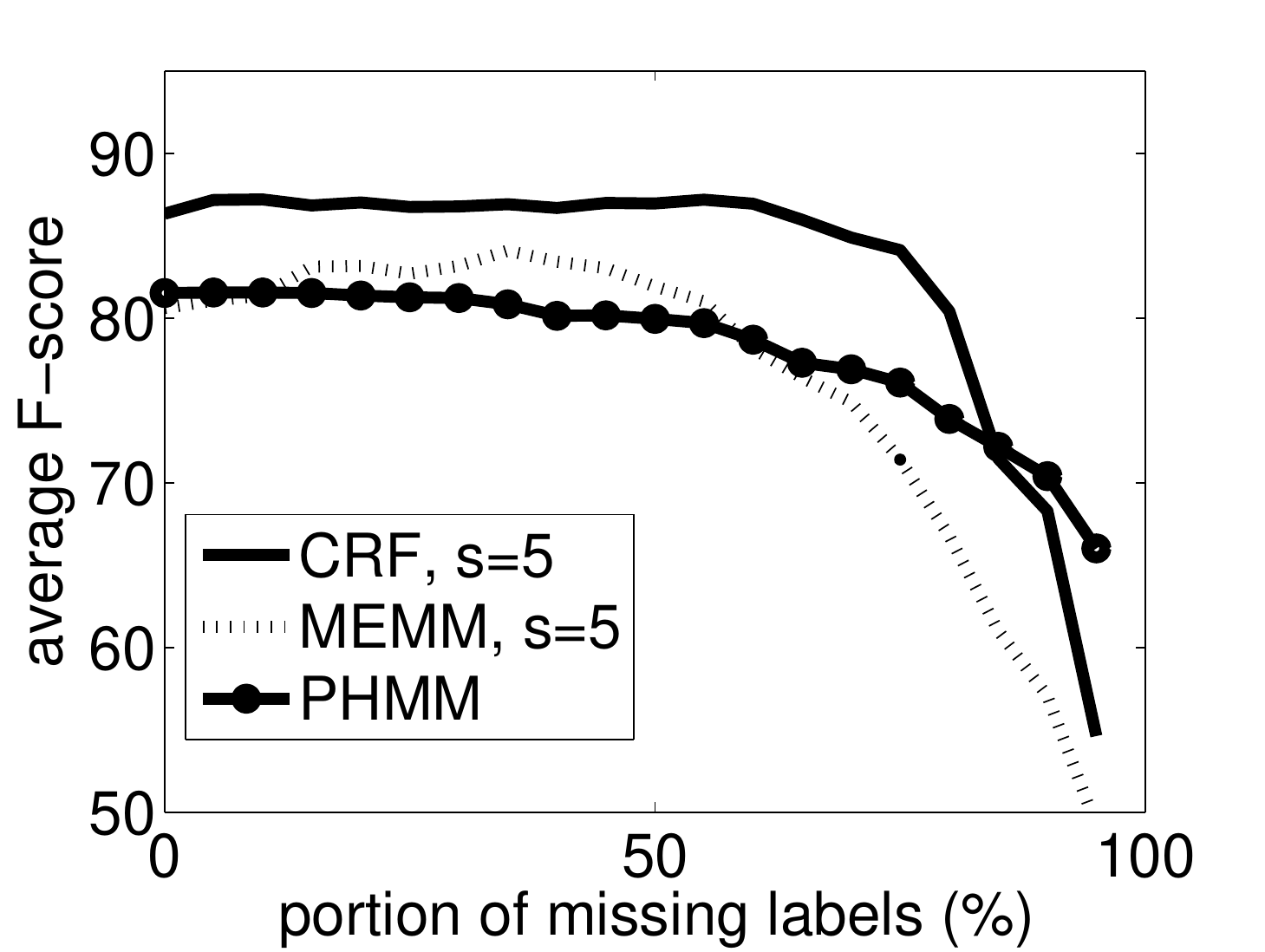}  & \includegraphics[width=0.3\linewidth]{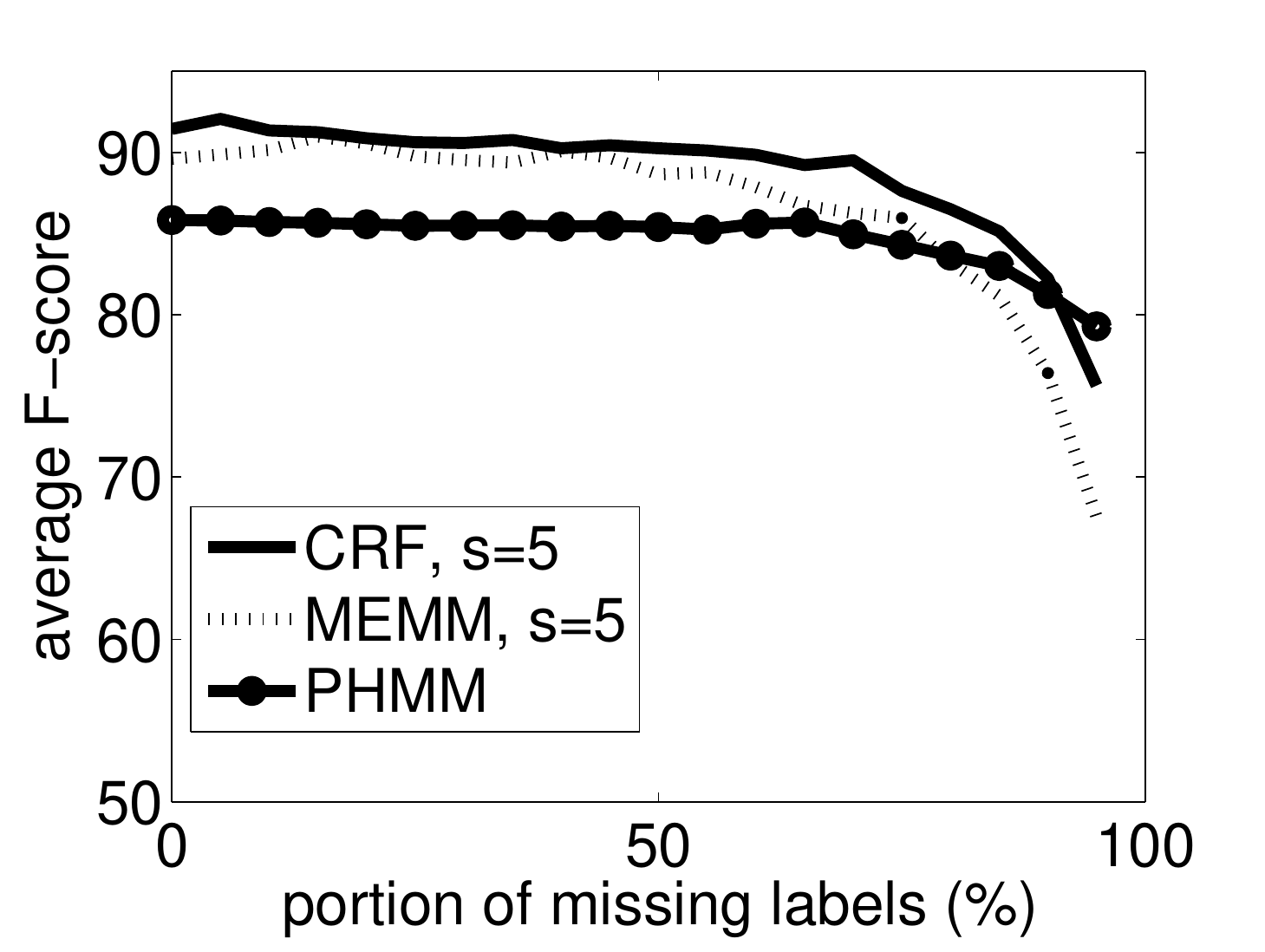}  & \includegraphics[width=0.3\linewidth]{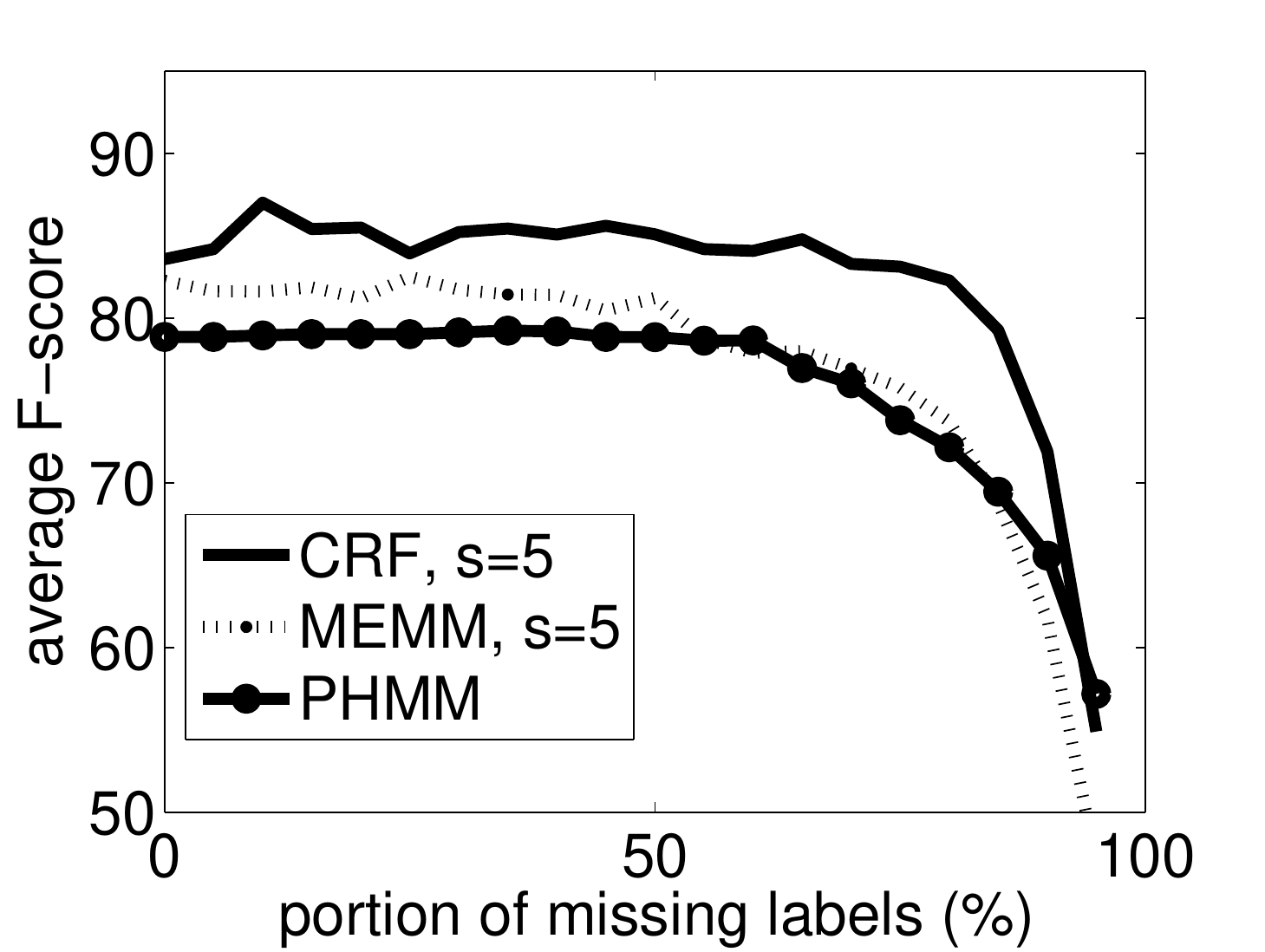} \tabularnewline
(a)  & (b)  & (c) \tabularnewline
\end{tabular}\caption{Average performance of models (a: SHORT\_MEAL, b: HAVE\_SNACK, c:
NORMAL\_MEAL). x-axis: portion of missing labels (\%) and y-axis:
the averaged F-score (\%) over all states and 10 repetitions.}

\par\end{centering}

\centering{}\label{fig:compare-all} 
\end{figure}

\begin{table}
\caption{The averaged precision ($P$) and recall ($R$) over all labels and
over 10 repetitions. Top row contains missing portion $\rho$. The
three scenarios: SM=SHORT\_MEAL, HS=HAVE\_SNACK, NM=NORMAL\_MEAL.}

\begin{centering}
{\small }%
\begin{tabular}{|l||l|c|c|c|c|c|c|c|c|c|c|c|}
\hline 
{\small Data} & {\small Model} & {\small Metric} & {\small 0 } & {\small 10 } & {\small 20 } & {\small 30 } & {\small 40 } & {\small 50 } & {\small 60 } & {\small 70 } & {\small 80 } & {\small 90 }\tabularnewline
\hline 
\hline 
{\small SM} & {\small CRF} & {\small $P$} & {\small 86.6} & {\small 86.3} & {\small 88.1} & {\small 86.9} & {\small 87.0} & {\small 89.9} & {\small 88.4} & {\small 83.8} & {\small 83.8} & {\small 72.5 }\tabularnewline
\hline 
{\small SM} & {\small CRF} & {\small $R$} & {\small 87.4} & {\small 87.1} & {\small 88.1} & {\small 87.7} & {\small 87.4} & {\small 91.3} & {\small 90.1} & {\small 81.6} & {\small 82.5} & {\small 68.5 }\tabularnewline
\hline 
{\small SM} & {\small MEMM} & {\small $P$} & {\small 81.7} & {\small 87.8} & {\small 87.0} & {\small 84.2} & {\small 85.2} & {\small 83.1} & {\small 81.2} & {\small 80.5} & {\small 73.2} & {\small 57.0 }\tabularnewline
\hline 
{\small SM} & {\small MEMM} & {\small $R$} & {\small 83.4} & {\small 88.4} & {\small 87.5} & {\small 84.2} & {\small 86.1} & {\small 82.7} & {\small 81.5} & {\small 75.8} & {\small 67.8} & {\small 55.2 }\tabularnewline
\hline 
{\small SM} & {\small HMM} & {\small $P$} & {\small 82.3} & {\small 82.3} & {\small 82.3} & {\small 81.1} & {\small 81.2} & {\small 80.8} & {\small 81.2} & {\small 79.9} & {\small 73.4} & {\small 66.9 }\tabularnewline
\hline 
{\small SM} & {\small HMM} & {\small $R$} & {\small 83.2} & {\small 83.2} & {\small 83.2} & {\small 83.7} & {\small 84.1} & {\small 83.3} & {\small 84.1} & {\small 83.1} & {\small 75.7} & {\small 70.9 }\tabularnewline
\hline 
{\small HS} & {\small CRF} & {\small $P$} & {\small 91.4} & {\small 90.4} & {\small 90.6} & {\small 91.5} & {\small 92.1} & {\small 89.7} & {\small 91.3} & {\small 91.5} & {\small 90.7} & {\small 89.5 }\tabularnewline
\hline 
{\small HS} & {\small CRF} & {\small $R$} & {\small 92.4} & {\small 91.5} & {\small 90.1} & {\small 90.6} & {\small 91.7} & {\small 90.0} & {\small 91.9} & {\small 91.5} & {\small 91.1} & {\small 88.8 }\tabularnewline
\hline 
{\small HS} & {\small MEMM} & {\small $P$} & {\small 89.9} & {\small 88.9} & {\small 90.8} & {\small 89.2} & {\small 91.5} & {\small 88.7} & {\small 89.6} & {\small 89.4} & {\small 85.1} & {\small 80.0 }\tabularnewline
\hline 
{\small HS} & {\small MEMM} & {\small $R$} & {\small 91.2} & {\small 90.3} & {\small 91.4} & {\small 89.5} & {\small 93.7} & {\small 90.4} & {\small 91.3} & {\small 91.0} & {\small 87.7} & {\small 81.4 }\tabularnewline
\hline 
{\small HS} & {\small HMM} & {\small $P$} & {\small 84.7} & {\small 84.7} & {\small 84.4} & {\small 85.0} & {\small 85.4} & {\small 85.3} & {\small 85.3} & {\small 85.3} & {\small 84.0} & {\small 79.4 }\tabularnewline
\hline 
{\small HS} & {\small HMM} & {\small $R$} & {\small 88.5} & {\small 88.5} & {\small 88.1} & {\small 87.2} & {\small 87.6} & {\small 87.3} & {\small 87.3} & {\small 87.3} & {\small 87.4} & {\small 83.4 }\tabularnewline
\hline 
{\small NM} & {\small CRF} & {\small $P$} & {\small 87.1} & {\small 88.9} & {\small 85.5} & {\small 83.7} & {\small 87.4} & {\small 85.4} & {\small 85.0} & {\small 86.8} & {\small 85.8} & {\small 74.0 }\tabularnewline
\hline 
{\small NM} & {\small CRF} & {\small $R$} & {\small 83.5} & {\small 88.5} & {\small 81.8} & {\small 80.7} & {\small 86.6} & {\small 85.7} & {\small 81.5} & {\small 86.3} & {\small 84.9} & {\small 72.8 }\tabularnewline
\hline 
{\small NM} & {\small MEMM} & {\small $P$} & {\small 85.4} & {\small 85.0} & {\small 84.6} & {\small 83.5} & {\small 84.8} & {\small 81.9} & {\small 77.9} & {\small 78.3} & {\small 75.0} & {\small 62.0 }\tabularnewline
\hline 
{\small NM} & {\small MEMM} & {\small $R$} & {\small 81.7} & {\small 82.1} & {\small 81.3} & {\small 81.0} & {\small 84.9} & {\small 81.4} & {\small 78.4} & {\small 79.7} & {\small 76.9} & {\small 62.6 }\tabularnewline
\hline 
{\small NM} & {\small HMM} & {\small $P$} & {\small 79.1} & {\small 79.1} & {\small 79.1} & {\small 79.1} & {\small 79.8} & {\small 79.8} & {\small 80.0} & {\small 77.1} & {\small 74.7} & {\small 58.3 }\tabularnewline
\hline 
{\small NM} & {\small HMM} & {\small $R$} & {\small 80.4} & {\small 80.4} & {\small 80.4} & {\small 80.4} & {\small 81.3} & {\small 81.3} & {\small 81.6} & {\small 79.5} & {\small 78.0} & {\small 63.8 }\tabularnewline
\hline 
\end{tabular}
\par\end{centering}{\small \par}

\label{precision-recall} 
\end{table}

Table \ref{precision-recall} and Figure \ref{fig:compare-all} show
performance metrics (precision, recall and $F1$-score) of all models
considered in this paper averaged over 10 repetitions. The three models
have equivalent graphical structures. The CRFs and MEMMs share the
same feature set but different from that of PHMMs. The generative
PHMMs are outperformed by the discriminative counterparts in all cases
given sufficient labels. This clearly matches the theoretical differences
between these models in that when there are enough labels, richer
information can be extracted in the discriminative framework, i.e.
modeling $p(y|x)$ is more suitable. On the other hand, when only
a few labels are available, the unlabeled data is important so it
makes sense to model and optimise $p(x,y)$ as in the generative framework.
On all data sets, the CRFs outperform the other models. These behaviours
are consistent with the results reported in \cite{lafferty01conditional}
in the fully observed setting. MEMMs are known to suffer from the
label-bias problem \cite{lafferty01conditional}, thus their performance
does not match that of CRFs, although MEMMs are better than HMMs given
enough training labels. In the HAVE\_SNACK data set, the performance
of MEMMs is surprisingly good.

A striking fact about the globally normalised CRFs is that the performance
persists until most labels are missing. This is clearly a big time
and effort saving for the labeling task. 

%% file: harem05_con.tex

\section{Conclusions and further work}

\label{sec:conclusion}

In this work, we have presented a semi-supervised framework for activity
recognition on low-level noisy data from sensors using discriminative
models. We illustrated the appropriateness of the discriminative models
for segmentation of surveillance video into sub-activities. As more
flexible information can be encoded using feature functions, the discriminative
models can perform significantly better than the equivalent generative
HMMs even when a large portion of the labels are missing. CRFs appear
to be a promising model as the experiments show that they consistently
outperform other models in all three data sets. Although less expressive
than CRFs, MEMMs are still an important class of models as they enjoy
the flexibility of the discriminative framework and enable online
recognition as in directed graphical models.

Our study shows that primitive and intuitive features work well in
the area of video surveillance. Semantically-rich and more discriminative
contextual features can be realised through the technique of a sliding
window. The wide context is especially suitable for the current problem
because human activities are clearly correlated in time and space.
However, to obtain the optimal context and to make use of the all
information embedded in the whole observation sequence, a feature
selection mechanism remains to be designed in conjunction with the
models and training algorithms presented in this paper.

Although flat CRFs and MEMMs can represent arbitrarily high-level
of activities, in many situations it may be more appropriate to structure
the activity semantics into multiple layers or into a hierarchy. Future
work will include models such as Dynamic Conditional Random Fields
(DCRFs) \cite{Sutton-et-al04}, conditionally trained Dynamic Bayesian
Networks and hierarchical model structures. A drawback of the log-linear
models considered here is the slow learning curve compared to the
traditional EM algorithm in Bayesian networks. It is therefore important
to investigate more efficient training algorithms.

\section*{Acknowledgments}

Hung Bui is supported by the Defense Advanced Research Projects Agency
(DARPA), through the Department of Interior, NBC, Acquisition Services
Division, under Contract No. NBCHD030010.

The authors would like to thank reviewers for suggestions to improve
the paper's presentation. The Matlab code of the L-BFGS algorithm
and of the conjugate gradient algorithm of Polak-Ribière is adapted
from S. Ulbrich and C. E. Rasmussen, respectively. The implementation
of PHMMs is based on the HMMs code by Sam Roweis.

%% file: harem05.bbl
\begin{thebibliography}{10}

\bibitem{aggarwal99human}
J.~K. Aggarwal and Q.~Cai.
\newblock Human motion analysis: {A} review.
\newblock {\em Computer Vision and Image Understanding: CVIU}, 73(3):428--440,
  1999.

\bibitem{Bui-et-al02}
H.~H. Bui, S.~Venkatesh, and G.~West.
\newblock Policy recognition in the abstract hidden {M}arkov model.
\newblock {\em Journal of Artificial Intelligence Research}, 17:451--499, 2002.

\bibitem{GRZEGORZ_EL-03}
Grzegorz Cielniak, Maren Bennewitz, and Wolfram Burgard.
\newblock Where is ...? {L}earning and utilizing motion patterns of persons
  with mobile robots.
\newblock In {\em Proceedings of the 18th International Joint Conference on
  Artificial Intelligence ({IJCAI})}, pages 909--914, Acapulco, Mexico, August
  2003.

\bibitem{Culotta-McCallum-AAAI05}
Aron Culotta and Andrew McCallum.
\newblock Reducing labeling effort for structured prediction tasks.
\newblock In {\em Proceedings of the National Conference on Artificial
  Intelligence (AAAI)}, 2005.

\bibitem{Dempster-et-al77}
A.~Dempster, N.~Laird, and D.~Rubin.
\newblock Maximum likelihood from incomplete data via the {EM} algorithm.
\newblock {\em Journal of Royal Statistical Society}, 39(1):1--38, 1977.

\bibitem{fine98hierarchical}
Shai Fine, Yoram Singer, and Naftali Tishby.
\newblock The hierarchical hidden {M}arkov model: Analysis and applications.
\newblock {\em Machine Learning}, 32(1):41--62, 1998.

\bibitem{Ivanov:00}
Y.~Ivanov and A.~Bobick.
\newblock Recognition of visual activities and interactions by stochastic
  parsing.
\newblock {\em IEEE Transactions on Pattern Analysis and Machine Intelligence
  (PAMI)}, 22(8):852--872, August 2000.

\bibitem{Kristjannson-et-al04}
Trausti Kristjannson, Aron Culotta, Paul Viola, and Andrew McCallum.
\newblock Interactive information extraction with constrained conditional
  random fields.
\newblock In {\em Proceedings of the 19th National Conference on Artificial
  Intelligence (AAAI)}, pages 412--418, San Jose, CA, 2004.

\bibitem{lafferty01conditional}
J.~Lafferty, A.~McCallum, and F.~Pereira.
\newblock Conditional random fields: Probabilistic models for segmenting and
  labeling sequence data.
\newblock In {\em Proceedings of the International Conference on Machine
  learning (ICML)}, pages 282--289, 2001.

\bibitem{liao2007learning}
Lin Liao, Donald~J Patterson, Dieter Fox, and Henry Kautz.
\newblock Learning and inferring transportation routines.
\newblock {\em Artificial Intelligence}, 171(5):311--331, 2007.

\bibitem{mcallum00maximum}
Andrew McCallum, Dayne Freitag, and Fernando Pereira.
\newblock {M}aximum {E}ntropy {M}arkov models for information extraction and
  segmentation.
\newblock In {\em Proceedings of the 17th International Conference on on
  Machine Learning (ICML)}, pages 591--598. Morgan Kaufmann, San Francisco, CA,
  2000.

\bibitem{OSENTOSKI_EL-04}
Sarah Osentoski, Victoria Manfredi, and Sridhar Mahadevan.
\newblock Learning hierarchical models of activity.
\newblock In {\em Proceedings of the IEEE/RSJ International Conference on
  Robots and Systems (IROS)}, 2004.

\bibitem{NIPS2005_810}
Ariadna Quattoni, Michael Collins, and Trevor Darrell.
\newblock Conditional random fields for object recognition.
\newblock In Lawrence~K. Saul, Yair Weiss, and {L\'{e}on} Bottou, editors, {\em
  Advances in Neural Information Processing Systems 17}, pages 1097--1104. MIT
  Press, Cambridge, MA, 2005.

\bibitem{RabinerIEEE-89}
Lawrence~R. Rabiner.
\newblock A tutorial on hidden {M}arkov models and selected applications in
  speech recognition.
\newblock {\em Proceedings of the {IEEE}}, 77(2):257--286, 1989.

\bibitem{Scheffer-Wrobel01}
T.~Scheffer and S.~Wrobel.
\newblock Active learning of partially hidden {M}arkov models.
\newblock In {\em Active Learning, Database Sampling, Experimental Design:
  Views on Instance Selection, Workshop at ECML-2001/PKDD-2001}, 2001.

\bibitem{sha-pereira:2003:HLTNAACL}
Fei Sha and Fernando Pereira.
\newblock Shallow parsing with conditional random fields.
\newblock In Marti Hearst and Mari Ostendorf, editors, {\em Proceedings of
  Human Language Technology (NAACL)}, pages 213--220, Edmonton, Alberta,
  Canada, May 27 - June 1 2003. Association for Computational Linguistics.

\bibitem{Sutton-et-al04}
C.A. Sutton, K.~Rohanimanesh, and A.~McCallum.
\newblock Dynamic {C}onditional {R}andom {F}ields: factorized probabilistic
  models for labeling and segmenting sequence data.
\newblock In {\em Proceedings of the International Conference on Machine
  learning (ICML)}, 2004.

\bibitem{Yamato-et-alCVPR92}
J.~Yamato, J.~Ohya, and K.~Ishii.
\newblock Recognizing human action in time-sequential images using hidden
  {M}arkov models.
\newblock In {\em Proceedings of the IEEE Conference on Computer Vision and
  Pattern Recognition (CVPR)}, pages 379--385, 1992.

\end{thebibliography}
